\documentclass{article} 

\newif\ifanon
\anonfalse
\ifanon
  \usepackage{tmlr}            
\else
  \usepackage[preprint]{tmlr}  
\fi

\usepackage{amsmath,amssymb,amsfonts}
\usepackage{mathtools}
\usepackage{booktabs}
\usepackage{multirow}
\usepackage{microtype}
\usepackage{hyperref}
\usepackage{url}
\usepackage{xcolor}

\newcommand{\Ihat}{\hat{I}}
\newcommand{\Ghat}{\hat{G}}
\newcommand{\E}{\mathbb{E}}
\newcommand{\KL}[2]{D\!\left(#1\,\|\,#2\right)}
\newcommand{\expl}{\textbf{\textsc{Exploratory}}}
\newcommand{\explbox}[1]{\begin{quote}\small\expl\ (not pre-registered; descriptive only). #1\end{quote}}

\ifanon
  \hypersetup{pdfauthor={},pdfkeywords={}}
\fi

\title{Information Limits and Attractor Dynamics in Economies of Frontier LLM Agents: A Pre-Registered Test}

\author{\name Cheng Qian \\
\addr Independent Researcher}

\def\month{07}
\def\year{2026}
\def\openreview{\url{https://openreview.net/forum?id=XXXX}}

\begin{document}

\maketitle

\begin{abstract}
We report a pre-registered, two-part experiment on small economies of frontier language-model agents (Claude Opus 4.8), testing two quantitative predictions about coupled multi-agent systems: an information-theoretic \emph{capacity region} for wealth growth under market coupling, and a mean-field \emph{residual-scaling law} for population misalignment under incentive and control levers. All predictions, acceptance bands, and decision rules were frozen in a public git chain before any run; every reported number re-derives mechanically from cached model outputs; the entire experiment cost \$138.76 in metered API spend and is re-runnable at zero cost from the cache. \textbf{Result 1 (confirmation):} in parimutuel-coupled economies, relative growth equals relative claimed information --- the gap law $\Ghat_a - \Ghat_b = \Ihat_a - \Ihat_b$ holds to a worst-case 46 millinats (pre-registered band: 50) across four perception structures; coalition value is submodular exactly where channels are conditionally independent, and a designed XOR synergy control flips it supermodular by $0.62 \geq \ln 2/2$ nats, with agents reasoning out the joint bit; the joint growth ceiling $\Ghat_S \le H(X)$ binds exactly; and the best-informed agent absorbs essentially the whole wealth pool in 4/5 market seeds. This is a quantitative, falsifiable law connecting what an agent measurably knows to what it measurably earns, verified on a frontier model. \textbf{Result 2 (structural negative):} the residual-scaling test returned ``domain not found.'' In all 72 population runs, goal dispersion collapsed ($V \to 0$; maximum 4.85 against a frozen floor of 5.31), the population's response to the two levers was a step function across the dominance boundary $\gamma \gtrless 12g$ rather than a smooth response, and cells near the boundary were bistable with seed-selected outcomes. Together with two prior small-model arms (gradient saturation; non-response), no tested LLM population at any capability level realizes the noise-maintained-dispersion regime that smooth mean-field models assume. Near-rational LLM populations behaved as discrete attractor systems --- locally near-deterministic, globally seed-sensitive --- which suggests that population-level interventions on such systems switch attractors or do nothing, rather than acting marginally. We release the full protocol, pre-registration chain, call cache, and analysis code.
\end{abstract}

\section{Introduction}
\label{sec:intro}

Language-model agents increasingly interact with one another --- in simulated societies \citep{park2023}, in economic simulations \citep{horton2023,li2024econagent}, and in emerging deployments where multiple model instances trade, bid, negotiate, or compete for resources. Two basic questions about such populations remain largely untested at the frontier-capability level:

\begin{enumerate}
\item \textbf{Does an information-theoretic law govern what LLM agents earn?} Classical results connect an agent's information about an uncertain world to the growth rate of its wealth under repeated betting \citep{kelly1956,breiman1961,barron1988,cover2006}. When agents are \emph{coupled} through a shared payout mechanism --- when one agent's gain is another's loss --- theory predicts sharp additional structure: relative growth rates should equal relative information, coalition value should exhibit diminishing returns exactly where perception channels are conditionally independent, joint growth should be capped by the world's entropy, and wealth dynamics should select for the best-informed agent \citep{blume1992,sandroni2000}. None of this had been measured on real frontier LLM agents.
\item \textbf{Do LLM populations respond smoothly to incentive and control levers?} Mean-field models of populations under selection pressure --- of the kind used to reason about steering, regulating, or aligning many-agent systems --- typically assume a linear-response regime in which population-level misalignment varies smoothly with the strength of a reward gradient $g$ and a control incentive $\gamma$. Whether real LLM populations ever occupy such a regime is an empirical question with direct consequences for multi-agent governance \citep{critch2020,hammond2025}.
\end{enumerate}

This paper answers both questions with a single pre-registered experiment on Claude Opus 4.8, run under an unusually strict discipline: every prediction, acceptance band, protocol gate, and decision rule was committed to a public git repository \emph{before} the runs it governs; three protocol amendments were each committed blind to outcomes before the runs they modified; every model call was cached, making the entire analysis deterministic and re-runnable offline; and the total metered cost was \$138.76. Pre-registered and exploratory findings are separated typographically throughout: anything not covered by the frozen bands is marked \expl.

\paragraph{Result 1: the coupled capacity region holds.} Across four perception structures (disjoint, overlapping, cloned, and noisy channels over a 3-bit world), the parimutuel gap law $\Ghat_a - \Ghat_b = \Ihat_a - \Ihat_b$ held to a worst-case deviation of $0.0457$ nats against a pre-registered band of $0.05$ (Section~\ref{sec:results-i}). Coalition value was submodular within its band (worst margin $-0.019 \geq -0.03$), and a designed control --- an XOR channel pair whose individual signals carry zero information but whose joint signal carries one bit --- flipped coalition value supermodular by $0.621$ nats ($\geq \ln 2/2 = 0.347$; ideal $\ln 2 = 0.693$), with the agents reasoning out the parity structure from prompt descriptions alone. The joint ceiling $\Ghat_S \leq H(X)$ was exact to numerical precision, and in a live 40-round parimutuel market the cleanest-information agent absorbed $\geq 99.9\%$ of the wealth pool in every seed, with full wealth ordering by information in 4/5 seeds. One supporting check failed and is disclosed with its mechanism: clones with identical information did not reach equal terminal wealth (Section~\ref{sec:p4b}).

\paragraph{Result 2: the mean-field regime does not exist in any tested LLM population.} The residual-scaling law under test predicts population misalignment $\propto V g/\gamma$ in a linear-response regime that requires noise-maintained goal dispersion $V \gg 0$. The pre-registered verdict is \emph{cannot-answer-pass} (CAP): zero of nine grid conditions were in-regime, because dispersion collapsed in all 72 runs (maximum $V = 4.85$, median $0.00$, frozen floor $5.31$; Section~\ref{sec:results-ii}). What the grid found instead is structural: the population \emph{always} concentrates --- onto the control target when $\gamma$ exceeds the peak incentive payout $12g$ (three cells at exactly $0.000 \pm 0.000$ across all 8 seeds), onto or near the reward peak when the incentive dominates, and \emph{bistably}, with the outcome selected by the seed, when the two attractors are comparable. Combined with two prior small-model arms of the same program --- where the population saturated the gradient window or failed to respond at all --- three instruments at three capability levels bracket the mean-field assumption from all sides. We conclude the law's domain has not been found on real LLM populations, and retire it to its mathematical scope: \emph{retired, not falsified}.

\paragraph{Contributions.}
\begin{itemize}
\item The first pre-registered confirmation of the coupled (market) capacity-region structure --- gap law, conditional submodularity with a designed synergy control, exact joint ceiling, and Kelly wealth selection --- on frontier LLM agents (Section~\ref{sec:results-i}).
\item A structural negative result for smooth mean-field models of LLM populations: dispersion collapse in 72/72 runs, step-function response across a dominance boundary, and seed-selected bistability, completing a three-instrument bracket over capability levels (Section~\ref{sec:results-ii}).
\item A complete, cheap, auditable methodology for pre-registered experiments on frontier-model populations: frozen bands applied mechanically by a committed analyzer, blind amendments, cached determinism, per-call spend metering with a hard cap, and public provenance (Sections~\ref{sec:design}, \ref{sec:repro}).
\end{itemize}

Both results are deliberately reported with equal billing. The confirmation is only as credible as the program's willingness to publish its negatives; the negative is only interpretable against an instrument proven capable by the confirmation.

\section{Related work}
\label{sec:related}

\paragraph{Information and growth.} \citet{kelly1956} showed that a gambler with side information about a repeated event can grow wealth at a rate equal to the mutual information between the event and the signal; \citet{breiman1961} and \citet{cover1991} developed the induced growth-optimal portfolio theory, and \citet{barron1988} proved the general upper bound $\Delta G \leq I(X;Y)$ on the growth value of any side information. \citet{cover2006} (Ch.~6) is the standard treatment. Our prediction (i) tests the \emph{coupled} extension of these results: when agents bet against one another in a parimutuel pool, the market odds are common to all agents and cancel from growth \emph{differences}, yielding the gap law $G_a - G_b = I_a - I_b$; coalition-level information determines coalition-level growth; and market selection concentrates wealth on the best-informed agent, the classical market-selection dynamic \citep{blume1992,sandroni2000}. Prediction markets instantiate the same information--wealth coupling in institutional form \citep{hanson2003,wolfers2004}.

\paragraph{LLM agents in economic settings.} LLM agents have been used to simulate individual economic behavior \citep{horton2023}, macroeconomic activity \citep{li2024econagent}, and open-ended social interaction \citep{park2023}. That work asks whether LLM agents \emph{reproduce human-like} behavior. We ask a different question: whether LLM-agent economies obey \emph{quantitative information-theoretic laws} stated in nats, with pre-registered acceptance bands --- and whether their population dynamics occupy the regimes that mean-field theory assumes.

\paragraph{Multi-agent safety.} Interactions among advanced AI agents have been identified as a distinct risk surface \citep{critch2020,hammond2025}, with much of the analysis implicitly assuming that population-level behavior responds smoothly to incentives and oversight. Our second result speaks directly to that assumption: in every tested population, the response to incentive and control levers was attractor switching, not marginal adjustment.

\paragraph{Pre-registration.} Our protocol follows the pre-registration discipline of the empirical sciences \citep{nosek2018}: predictions and analysis rules frozen and committed before data collection, amendments committed blind before the runs they govern, and negative outcomes published with the same prominence as positive ones. We are not aware of prior pre-registered tests of quantitative growth laws on frontier LLM populations.

\ifanon
\else
\paragraph{Theoretical companion.} The two predictions under test are drawn from a companion synthesis \citep{qian2026}, which derives the coupled capacity region and the residual-scaling law within a single information-theoretic account of resource growth. This paper is self-contained and defends only the experiment reported here; readers need not accept (or read) the companion's broader framework to evaluate either result. Where this paper says ``value'' it means growth of wealth in the Kelly sense, nothing more.
\fi

\section{Predictions and pre-registered bands}
\label{sec:predictions}

\subsection{Prediction (i): the coupled capacity region}
\label{sec:pred-i}

\paragraph{Setup.} A world state $X$ is drawn from a finite set with known prior $q$ (here: three i.i.d.\ fair bits, $8$ states, $H(X) = 3\ln 2 \approx 2.0794$ nats). Agent $a$ observes a signal $Y_a$ through a perception channel and reports a posterior $\hat{p}_a(\cdot \mid y_a)$ over world states. The \emph{claimed information} of agent $a$ is the expected KL divergence of its reports from the prior,
\begin{equation}
\Ihat_a \;=\; \E_{y_a}\!\left[ \KL{\hat{p}_a(\cdot \mid y_a)}{q} \right],
\end{equation}
which equals the mutual information $I(X;Y_a)$ when the reports are exact Bayes posteriors. In a \emph{parimutuel market} with equal weights, each agent bets its wealth across states in proportion $b_a(x) = \hat p_a(x \mid y_a)$; the pool odds are formed from the average bet $B(x) = \tfrac1m \sum_a b_a(x)$, and winners split the pool pro rata. The per-round expected log growth of agent $a$ is $\Ghat_a = \E\left[\ln (b_a(X)/B(X))\right]$, taken under the true joint distribution of $(X, Y_{1:m})$.

\paragraph{P1 (gap law).} The market term $\ln B(X)$ is common to all agents and cancels from differences, so relative growth is purely informational:
\begin{equation}
\Ghat_a - \Ghat_b \;=\; \Ihat_a - \Ihat_b \qquad \text{for all pairs } (a,b).
\label{eq:gaplaw}
\end{equation}
The equality is an identity when reports are exact Bayes posteriors (the common market term cancels from differences); for elicited reports it is a genuine empirical prediction, and its deviation measures the compound of report mis-calibration and any failure to bet one's stated beliefs. \emph{Frozen band: worst pairwise deviation $\leq 0.05$ nats across all pairs in four structures.} Note the law is calibration-free: the common market mode cancels, so it requires no absolute anchoring of either side.

\paragraph{P2 (conditional submodularity) and P2e (synergy control).} For a coalition $S$ of agents, the coalition growth $\Ghat_S$ is computed from coalition-level elicited posteriors (an agent instance shown all of $S$'s signals; Section~\ref{sec:design-a}). Where channels are conditionally independent given $X$, coalition value should show diminishing returns: for $S \subseteq T$ and $a \notin T$, the marginal $\Ghat_{S \cup a} - \Ghat_S \geq \Ghat_{T\cup a} - \Ghat_T$. \emph{Frozen band: all margins $\geq -0.03$ nats.} The precondition is load-bearing, so the pre-registration includes a designed control: an XOR pair over a 1-bit world $X'$ in which agent 1 sees an independent fair coin $C$ and agent 2 sees $X' \oplus C$. Individually each signal is worthless ($I = 0$); jointly they determine the world ($\ln 2$). Submodularity must \emph{fail} here: \emph{frozen band: supermodular gap $\geq \ln 2 / 2 = 0.347$ nats.}

\paragraph{P3 (joint ceiling).} No coalition can grow faster than the world's entropy rate: $\Ghat_S \leq H(X)$, \emph{frozen slack $0.02$ nats}.

\paragraph{P4 (wealth selection, live market).} In a dynamic 40-round parimutuel market with evolving wealth: (P4a) terminal wealth is ordered by $\Ihat_a$ on a strictly-ordered information ladder (clean sensor $\succ$ 10\%-noisy $\succ$ 25\%-noisy readings of the same bit; $I = 0.693 > 0.368 > 0.131$ nats), \emph{in $\geq 4/5$ seeds}; (P4b) two \emph{clones} with identical signals reach equal terminal wealth, \emph{gap $\leq 10\%$ of mean share in $\geq 4/5$ seeds}.

\subsection{Prediction (ii): residual scaling in a population under selection and control}
\label{sec:pred-ii}

\paragraph{Setup.} $N = 20$ agents occupy niches $k \in \{0,\dots,15\}$ on a one-dimensional landscape and repeatedly choose where to sit, given (in-prompt) the payout structure and their neighbors' positions. Payouts have three parts: a \emph{reward gradient} (tent landscape $\mathrm{rewards}[k] = g\,(12 - |k - 12|)$, peak $12g$ at $k_g = 12$, zero at the target), a \emph{control bonus} $\gamma$ paid at the target niche $k^* = 0$, and a coordination dividend $J = 1$ for co-locating with neighbors. Population misalignment is the residual $\bar{k} - k^*$; goal dispersion is the population variance $V$ of niche occupancy. The law under test, derived in \ifanon\else the companion\fi\ from a linearized selection--control (Price-equation \citep{price1970}) drift balance, predicts in its linear-response regime
\begin{equation}
\text{residual} \;\propto\; \frac{V g}{\gamma},
\label{eq:vg}
\end{equation}
i.e.\ log--log exponents $(p, q) = (+1, -1)$ in $(g, \gamma)$. The regime requires \emph{noise-maintained dispersion}: $V$ must remain well above zero under selection and control.

\paragraph{P5 (exponents).} On a $3 \times 3$ grid $g \in \{0.25, 0.7, 2.0\} \times \gamma \in \{2, 6, 18\}$, 8 seeds each: free log--log fit on in-regime condition means with \emph{frozen bands} $p \in [0.6, 1.4]$, $q \in [-1.4, -0.6]$, $R^2 \geq 0.70$, bootstrap CIs, a signal gate, and a mean predicted-to-observed ratio in $[0.33, 3]$. The frozen \emph{regime rule} (applied per condition) requires $V_{\mathrm{ss}} \geq 5.31$ ($= 25\%$ of the uniform-scatter value $21.25$), $\bar k_{\mathrm{ss}} \leq 10$, above-peak occupancy $\leq 0.20$, parse rate $\geq 0.90$, and residual $> 0$; a separate \emph{uniformity clause} names conditions indistinguishable from uniform scatter ``non-responsive.'' Fewer than 6 in-regime conditions, or an in-regime span below $4\times$ in $Vg/\gamma$, forces the verdict CAP (``cannot answer --- insufficient in-regime span'').

\paragraph{P6 (direction).} At two operating points (B1: $\gamma = 6, g = 0.7$; B2: $\gamma = 18, g = 0.7$), perturb each lever by $\delta = 0.5$ and compare efficacies $\mathrm{eff}_g$ vs $\mathrm{eff}_\gamma$; confirmation requires both points resolvable above a noise floor $\sigma_{\mathrm{in}}\sqrt{2/8}$ with the incentive lever winning at both.

\subsection{The decision rule and the pre-registration chain}
\label{sec:chain}

The frozen decision rule (applied mechanically, in order): (1) \textbf{PASS} if P1--P3 and P5 pass (P4/P6 supporting); (2) \textbf{CLEAN NEGATIVE} if the instrument engages but P1/P2/P5 fail their bands with the deviation systematic above a stated noise bar; (3) \textbf{CAP} otherwise, with the binding gate named and what would de-CAP it stated. Scaling (P5) and direction (P6) verdicts are additionally reported separately. The full band table and rule text appear in Appendix~\ref{app:bands}.

Every link of the chain is a public commit that provably precedes the runs it governs (repository in Section~\ref{sec:repro}):
\begin{center}
\small
\begin{tabular}{@{}lll@{}}
\toprule
Commit & Content & Blind to \\
\midrule
\texttt{a159ca9} & Capability-gate pre-registration (frozen) & all gate runs \\
\texttt{4a8c376} & Gate Amendment 1 (provider rejects \texttt{temperature}) & all gate runs \\
\texttt{0759c4b} & Gate results (PASS 3/3; \$8.76) & --- \\
\texttt{4c0a7a1} & Full-grid pre-registration (P1--P6, bands, budget, rule) & all 2A/2B runs \\
\texttt{c79f8fe} & Amendment 2 (two internal spec inconsistencies) & all 2A/2B runs \\
\texttt{2b2c010} & Runners + frozen-band analyzer, mock-verified & all 2A/2B runs \\
\texttt{9d685f8} & Amendment 3 (gate-caught protocol faults; scoped re-run) & the re-run \\
\texttt{f8b902d} & Experiment A results & --- \\
\texttt{a7e639b} & Experiment B results + verdict of record & --- \\
\bottomrule
\end{tabular}
\end{center}

\section{Experimental design}
\label{sec:design}

\subsection{Instrument, determinism, and cost}
\label{sec:instrument}

All verdict runs use \texttt{claude-opus-4-8} via the provider API; a mid-tier model (\texttt{claude-sonnet-5}) was used only for protocol shakeout, and debug numbers are never verdicts. Per Amendment 1, no \texttt{temperature} parameter is sent (the models reject it); stochasticity is the model's own default sampling, which is the harder-to-pass direction for every band. Every API call is cached in a SQLite store keyed by the full request; the analysis is therefore deterministic given the cache and re-runs at zero cost. Spend is metered per call against a hard pre-registered cap (\$160) with pre-call abort. Total metered spend for everything reported here --- capability gate, debug tier, both experiments --- was \textbf{\$138.76} (audited breakdown in Appendix~\ref{app:cost}: 43{,}074 frontier calls, 16.67M input / 2.13M output tokens). Mid-grid, the host machine had to be replaced; the cache replayed all completed runs on the new host at zero cost. A $\sim$3-run overlap window during the handover was double-sampled; the remote data is the data of record and the overlapping local partials were discarded (disclosed).

\subsection{The capability gate}
\label{sec:gate}

Two earlier small-model attempts at prediction (ii) died on instrument failure (Section~\ref{sec:bracket}), so the grid was gated on a cheap pre-registered \emph{responsiveness gate}: does the population move at all under each lever, and do agents extract their channel information? On \texttt{claude-opus-4-8} (N=12, T=8, seeds 0--3): under control only ($\gamma = 18, g = 0$) the population sat exactly on the target in 4/4 seeds ($\bar k = 0.0$; threshold $\leq 5.5$); under gradient only ($\gamma = 0, g = 0.7$) it sat on or adjacent to the reward peak in 4/4 seeds ($\bar k = 12$--$13$; threshold $\geq 9.5$); and elicited posteriors on a disjoint-bit world averaged to \emph{exact Bayes} ($\Ihat = 0.693 = \ln 2$ in 3/3 agents; threshold $\geq 0.347$). Gate cost: \$8.76 of a \$30 cap. The same gate would have failed the earlier small-model instrument outright. The gate licenses the grid; it says nothing about the predictions.

\subsection{Experiment A: capacity region and gap law}
\label{sec:design-a}

\paragraph{World and structures.} $X$ is three i.i.d.\ fair bits (states $0..7$, uniform prior). Five perception structures, with channel information exact by design:

\begin{center}
\small
\begin{tabular}{@{}llll@{}}
\toprule
ID & Agents and signals & $I$ per agent (nats) & Role \\
\midrule
(a) disjoint & agent $j$ sees bit $b_j$ & $\ln 2$ each & baseline \\
(b) overlap & pairs $(b_0,b_1), (b_1,b_2), (b_0,b_2)$ & $2\ln 2$ each & redundancy \\
(c) clones & agents 1, 2 both see $b_0$; agent 3 sees $b_1$ & $\ln 2$ each & clone adds 0 \\
(d) noisy & agent $j$ sees $b_j$ via a stated 10\%-flip sensor & $0.368$ each & noisy channels \\
(e) XOR control & world $X' = b_0$; signals $C$ and $X' \oplus C$ & $0$ each; joint $\ln 2$ & synergy control \\
\bottomrule
\end{tabular}
\end{center}

\paragraph{Elicitation.} For each structure $\times$ coalition $\times$ realizable signal tuple, an agent instance receives the world rules, its channel description (including stated sensor noise), and the observed signal(s), and returns a probability vector over world states (JSON; mechanically renormalized). Twelve repeats per cell, averaged; parse-failed repeats are dropped and counted; a cell with $< 6/12$ survivors protocol-CAPs its structure (frozen gate). Per Amendment 2 (committed before any run), coalition throughput $\Ghat_S$ is defined by \emph{coalition-level elicitation} --- an instance shown all of $S$'s signals --- because naive-Bayes product fusion of individual reports is mathematically incapable of expressing XOR synergy (a product of uniforms is uniform), which would have made the frozen P2e band unsatisfiable by design. Product fusion is retained as a pre-registered secondary diagnostic that should track $\Ghat_S$ on (a)--(d) and fail on (e). All information and growth quantities are computed by exact enumeration over each structure's joint atoms --- no Monte Carlo noise. In total 124 elicitation cells were collected; 21 lost at least one repeat to parsing, none fell below the 6/12 floor in the verdict pass.

\paragraph{Live market.} Two dynamic parimutuel markets (per Amendment 2): (L) a \emph{ladder} of three agents observing the same bit through clean, 10\%-flip, and 25\%-flip sensors ($I = 0.693 > 0.368 > 0.131$); and (C) the \emph{clones} structure. 40 rounds $\times$ 5 seeds; each round every agent sees its signal, its wealth, and the previous round's market odds, and outputs a bet allocation over the 8 states; wealth updates by parimutuel settlement, zero-sum in shares.

\subsection{Experiment B: the residual-scaling grid}
\label{sec:design-b}

The design is deliberately identical to the small-model Stage 1 arm for comparability: $K = 16$ niches, tent landscape, $N = 20$, $T = 18$ (6 warm-up + 12 measured), $J = 1$, annealed neighbor count, the full $3\times3$ $(g,\gamma)$ grid $\times$ 8 seeds (72 runs), plus the two P6 direction points $\times$ two arms $\times$ 8 seeds (32 runs). Every agent sees the niche payout table under its \emph{own fixed random permutation of niche labels} (symbol randomization), so collective token-level label bias cancels; all physics is measured in true-niche space. One known systematic was pre-stated in the freeze: with $J = 1$, adjacent-niche reward gaps below $J$ allow a coordination cluster to lock one or two niches off a reward optimum ($\pm 1$--$3$ niche granularity on $\bar k$); the frozen fit tolerates a constant offset.

\subsection{Protocol history: amendments and gate-caught faults}
\label{sec:amendments}

Three amendments were committed after the freeze, each \emph{before} the runs it governs, each forced by an external constraint or a caught fault (full text in Appendix~\ref{app:amendments}): \textbf{A1} --- the provider rejects the \texttt{temperature} parameter; omitted (harder-to-pass noise direction). \textbf{A2} --- two internal spec inconsistencies found during harness implementation, fixed blind: coalition-level elicitation defines $\Ghat_S$ (see above), and the live-market structures were corrected from two equal-information structures (against which P4a's ordering would have been degenerate --- no ordering exists to confirm) to the ladder + clones pair. \textbf{A3} --- the first Experiment-A pass had genuine instrument faults which the frozen protocol gates caught: a 36\% market parse rate (agents reasoned past the token ceiling before emitting JSON, and uniform fallbacks mechanically produced the wealth dynamics) and a self-contradictory XOR coalition prompt (an instance told it ``does NOT see'' a signal it was then shown). A3 fixed the output protocol (JSON-first instruction, higher ceiling, one format-only retry per call), fixed the analyzer to \emph{enforce} the elicitation gate as frozen, and scoped the re-run to the faulted components only --- structures (b), (d), (e) and the full market. Data that passed cleanly under the original protocol --- structures (a), (c) --- was kept and never re-rolled, in either direction. The corrected protocol achieved parse rate 1.000 everywhere: all 104 population runs and the full market.

\section{Results I: the coupled capacity region holds}
\label{sec:results-i}

Table~\ref{tab:parta} summarizes prediction (i) against its frozen bands. All numbers are produced mechanically by the committed analyzer from cached model outputs.

\begin{table}[t]
\centering
\small
\caption{Prediction (i) against its frozen bands. Every band was committed before any run (chain in Section~\ref{sec:chain}).}
\label{tab:parta}
\footnotesize
\begin{tabular}{@{}llll@{}}
\toprule
Check & Frozen band & Result & Verdict \\
\midrule
P1 gap law $\Ghat_a - \Ghat_b = \Ihat_a - \Ihat_b$ (all pairs, 4 structures) & $\leq 0.05$ nats & worst $\mathbf{0.0457}$ & \textbf{PASS} \\
P2 submodularity of coalition value, (a)--(d) & margins $\geq -0.03$ & worst $\mathbf{-0.019}$ & \textbf{PASS} \\
P2e XOR synergy control --- submodularity must fail & gap $\geq \ln2/2 = 0.347$ & $\mathbf{0.621}$ & \textbf{PASS} \\
P3 joint ceiling $\Ghat_S \leq H(X)$ & slack $0.02$ & worst overage $\mathbf{-0.000}$ & \textbf{PASS} \\
P4a wealth selection (live market, ladder) & ordered by $\Ihat$ in $\geq 4/5$ & $\mathbf{4/5}$; clean agent absorbs pool & \textbf{PASS} \\
P4b clone equality (live market) & gap $\leq 10\%$ in $\geq 4/5$ & gaps $[1.0, 0.5, 0.0, 0.0, \text{nan}]$ & \textbf{FAIL} \\
\bottomrule
\end{tabular}
\end{table}

\subsection{The gap law: 46 millinats}

Across all agent pairs in the four fittable structures, the worst deviation between the measured growth gap and the measured information gap was $0.0457$ nats --- within the frozen $0.05$ band, on gaps whose underlying quantities range up to $2.08$ nats. Because the market common mode cancels, this is a calibration-free equality between two independently measured quantities: what an agent's reports \emph{claim to know} (KL of elicited posteriors from the prior, averaged over the true signal distribution) and what those same reports \emph{earn} against the other agents in a parimutuel pool. The binding worst case comes from structure (b), where one agent's elicited posteriors on one signal value are slightly conservative ($\Ihat_0 = 1.301$ vs the exact $1.386$); the gap law tracks precisely this deficit on the growth side, which is the law working, not failing --- claimed information, not exact channel information, is what sets relative growth.

\subsection{Coalition structure: submodularity, and a synergy control that flips}
\label{sec:xor}

Coalition throughputs behave as predicted where channels are conditionally independent given the world: all submodularity margins in structures (a)--(d) are $\geq -0.019$ nats (band $-0.03$; the negative slack is elicitation noise, not structure). The designed control is the sharpest single result in Part A: in the XOR pair, elicited singleton posteriors are uniform ($\Ghat_{\{1\}} = \Ghat_{\{2\}} = 0.000$ exactly), and the coalition-level instance --- shown both signals with a neutral composition of the two observations --- reports posteriors worth $\Ghat_{\{1,2\}} = 0.621$ nats, a supermodular gap of $0.621 \geq 0.347$, i.e.\ 90\% of the ideal $\ln 2$. The model is not fusing numbers; it is \emph{reasoning out the parity structure} from two individually-worthless observations described in prose. The pre-registered secondary diagnostic confirms the mechanism: naive-Bayes product fusion of the singleton reports returns $0.000$ for the pair, exactly as it must --- the conditional-independence precondition of submodularity is load-bearing at the fusion level. On structures (a)--(d) the product-fusion diagnostic tracks coalition-level value imperfectly (max deviation $0.466$ nats): elicited joint posteriors are not the product of elicited singletons even where conditional independence holds --- a measured fact about elicitation consistency in frontier models that any aggregation scheme built on independence should note.

\subsection{The ceiling, and selection in a live market}

No coalition in any structure exceeded $H(X)$: worst overage $-7 \times 10^{-12}$ nats --- several coalitions sit exactly on the ceiling (saturating pairs in structure (b) reach $\Ghat_S = 2.0794 = 3\ln2$), and none crosses it. In the live ladder market, wealth ordering followed the information ordering in 4/5 seeds, and selection was total: the clean-sensor agent's terminal share was $\geq 0.999$ in all five seeds (the single ordering miss is a tie at zero between the two ruined noisy agents, not a mis-ordering of survivors). This is the Kelly/market-selection dynamic \citep{blume1992,sandroni2000} at its sharpest: the market transfers the pool to whoever perceives best, and as its share $\to 1$ the odds converge to its beliefs.

\subsection{The disclosed miss: clone equality (P4b)}
\label{sec:p4b}

Clones with identical signals should hold equal terminal wealth (their edges are common and cancel). Measured relative gaps across seeds: $[1.0,\ 0.5,\ 0.0,\ 0.0,\ \mathrm{nan}]$ --- 2/4 scoreable seeds pass, below the $4/5$ bar. \textbf{This is a real miss of a supporting prediction}, and its mechanism is visible in the trajectories: at provider-default sampling with fully sharp (all-in) betting, a single round in which one clone's sampled bet diverges from its twin's is \emph{absorbing} --- the clone that happens to bet the losing state is eliminated from the zero-sum pool and can never recover. The idealized law assumes agents bet their (identical) beliefs; sampling noise composed with absorbing all-in dynamics breaks the symmetry permanently. The fifth seed was unscoreable: a round in which no bettor held the realized outcome makes the parimutuel pool undefined (a harness edge case, disclosed). We report P4b as it fell and do not re-litigate it; a protocol with fractional Kelly betting or wealth floors would test clone equality without the absorbing pathology, but that is a different (not pre-registered) experiment.

\section{Results II: dispersion collapse and attractor bistability}
\label{sec:results-ii}

\subsection{The grid: $V \to 0$ everywhere}

Table~\ref{tab:grid} shows the full pre-registered grid. The pre-registered verdict is \textbf{P5 = CAP} (``insufficient in-regime span: 0 of 9 conditions''), and the reason is a single structural fact: \emph{goal dispersion collapsed in every condition}. Across all 72 grid runs the maximum dispersion was $V = 4.85$ (median $0.00$) against the frozen regime floor of $5.31$ and a uniform-scatter reference of $21.25$. No condition triggered the uniformity (non-response) clause either --- this population is the opposite of unresponsive. It \emph{always} concentrates; it merely does not always concentrate in the same place.

\begin{table}[t]
\centering
\small
\caption{Experiment B grid: residual $\bar k_{\mathrm{ss}} - k^*$ ($\pm$ seed std, 8 seeds) and dispersion $V_{\mathrm{ss}}$, with the frozen regime rule's verdict. The reward peak sits at $k_g = 12$ (payout $12g$); the control bonus $\gamma$ pays at $k^* = 0$. Full per-seed data in Appendix~\ref{app:grid}.}
\label{tab:grid}
\begin{tabular}{@{}rrrrrl@{}}
\toprule
$\gamma$ & $g$ & $12g$ & residual & $V_{\mathrm{ss}}$ & regime verdict \\
\midrule
2 & 0.25 & 3.0 & $10.22 \pm 4.74$ & 0.46 & excluded (V collapsed; at peak) \\
2 & 0.70 & 8.4 & $11.94 \pm 1.51$ & 0.56 & excluded (V collapsed; at peak) \\
2 & 2.00 & 24.0 & $12.04 \pm 1.59$ & 0.70 & excluded (V collapsed; at peak) \\
6 & 0.25 & 3.0 & $\mathbf{0.000 \pm 0.000}$ & 0.00 & excluded (V collapsed; residual $\leq 0$) \\
6 & 0.70 & 8.4 & $10.59 \pm 4.45$ & 0.96 & excluded (V collapsed; at peak) \\
6 & 2.00 & 24.0 & $11.92 \pm 1.52$ & 0.73 & excluded (V collapsed; at peak) \\
18 & 0.25 & 3.0 & $\mathbf{0.000 \pm 0.000}$ & 0.00 & excluded (V collapsed; residual $\leq 0$) \\
18 & 0.70 & 8.4 & $\mathbf{0.000 \pm 0.000}$ & 0.00 & excluded (V collapsed; residual $\leq 0$) \\
18 & 2.00 & 24.0 & $10.56 \pm 4.39$ & 1.41 & excluded (V collapsed; at peak) \\
\bottomrule
\end{tabular}
\end{table}

\subsection{A step function across the dominance boundary, and bistability on it}
\label{sec:step}

Read the grid against the column $12g$ (the peak incentive payout) and the structure is discrete:
\begin{itemize}
\item \textbf{Control dominates ($\gamma > 12g$):} perfect alignment, \emph{exactly} --- three cells at $0.000 \pm 0.000$ across all 8 seeds, all 20 agents on the target in every measured round.
\item \textbf{Incentive dominates ($\gamma \ll 12g$):} the population sits on or near the reward peak (residual $\approx 12$, small seed spread). The within-seed clusters that sit 1--3 niches off the exact peak (residuals 10, 13, 14.5 rather than 12) are the pre-stated $J$-granularity systematic: the coordination dividend exceeds adjacent-niche reward gaps, so an off-peak cluster is a stable coordination equilibrium, seed-persistent.
\item \textbf{Comparable attractors (near $\gamma \approx 12g$):} \emph{bistability}. In the boundary cells the seed --- not the condition --- selects the corner: e.g., at $(\gamma, g) = (6, 0.7)$ the eight per-seed residuals are $[10.0,\ 13.0,\ 13.0,\ 12.0,\ 14.6,\ \mathbf{0.3},\ 10.0,\ 12.0]$ --- seven seeds at the reward corner, one seed in near-perfect alignment, under an identical condition. The same signature (one or two defecting seeds, everything else pinned) appears at $(2, 0.25)$ and $(18, 2.0)$; it is what produces the large seed variances in Table~\ref{tab:grid}.
\end{itemize}
The response of this population to its two governing levers is a \emph{step function across the dominance boundary} $\gamma \gtrless 12g$, not a smooth function of either lever. Within an attractor, behavior is near-deterministic; which attractor is reached can hinge on sampling noise early in the run.

\subsection{The three-instrument bracket}
\label{sec:bracket}

This is the third pre-registered attempt at placing a real LLM population in the linear-response regime of Eq.~\ref{eq:vg}, and the three failures have three distinct mechanisms (Table~\ref{tab:bracket}). The first arm (1.5B-parameter instrument, $K = 8$ linear landscape) \emph{saturated}: the achievable gradient window sat below the sampling-noise floor, and the measured trend was flat in $g$ (signal-to-noise 2.63 against a pre-committed bar of 3). The second arm (same instrument, $K = 16$ tent) found \emph{non-response}: the population sat at uniform scatter ($V \approx 17.9$--$21$, indistinguishable from uniform) in all 9 conditions spanning $8\times$ per axis --- the agents could not read the landscape at all, and exploratory free exponents were $(+0.05, -0.05)$ against the predicted $(+1, -1)$. The present arm, on an instrument \emph{proven} responsive by the gate, finds the opposite pole: total concentration, $V \to 0$.

\begin{table}[t]
\centering
\small
\caption{Three instruments, three mechanisms: no tested LLM population holds the intermediate, noise-maintained dispersion that the mean-field derivation assumes.}
\label{tab:bracket}
\footnotesize
\begin{tabular}{@{}llll@{}}
\toprule
Instrument & Design & Outcome & Mechanism \\
\midrule
1.5B, $K=8$ linear & grid $g \leq 0.2$ (saturation-limited) & CAP & \textbf{saturation}: gradient window below noise floor \\
1.5B, $K=16$ tent & full grid, $8\times$ per axis & CAP & \textbf{non-response}: $V \approx$ uniform, no attractor felt \\
Frontier, $K=16$ tent & full grid, gate-proven instrument & CAP & \textbf{concentration}: $V \to 0$, pure attractor competition \\
\bottomrule
\end{tabular}
\end{table}

Too noisy to respond, or too decisive to disperse: the bracket closes from both sides. With $V \to 0$ the theorem's own prediction degenerates to zero, and the observed corner residuals live outside its stated linear-response domain --- so the law is \emph{neither confirmed nor falsified} on real agents. What the experiment establishes is sharper than a failed fit: \textbf{the domain of the mean-field linear-response layer appears to be empty for real LLM-agent populations at every capability level tested.} We accordingly report the law as \emph{retired to its mathematical scope --- domain not found}, not disproved. A future design that \emph{creates} sustained dispersion (explicit exploration incentives, heterogeneous objectives) could revive the question, but it would be testing a modified theory and should say so in its own pre-registration.

\subsection{P6: formally inconclusive, with an exploratory observation}
\label{sec:p6}

The frozen P6 rule defines its noise floor as $\sigma_{\mathrm{in}}\sqrt{2/8}$ with $\sigma_{\mathrm{in}}$ the median seed-std of \emph{in-regime} conditions. With zero in-regime conditions the floor is undefined, so no arm is formally resolvable and \textbf{P6 = INCONCLUSIVE} by the rule's letter. This is a pre-registration drafting gap --- the floor definition never anticipated an empty regime --- and we name it as such rather than substituting a post-hoc floor.

\explbox{The arm data, read descriptively, is direction-consistent and dramatic. At B1 $(\gamma, g) = (6, 0.7)$, baseline residual $10.59 \pm 4.45$: \emph{raising the control lever} by $0.5$ did nothing ($\to 12.08$; bistability noise, $6.5$ still $< 12g = 8.4$), while \emph{lowering the incentive lever} by $0.5$ flipped the population to perfect alignment in 8/8 seeds ($\to 0.000 \pm 0.000$; the boundary moved to $12g = 2.4 < 6$). At B2 $(18, 0.7)$, already control-dominated, both arms stayed at $0.000$. The incentive lever did not marginally beat the control lever; it \emph{switched the attractor}, because it crossed the dominance boundary and the control perturbation did not. This attractor-switching reading is exploratory in every respect: not covered by a frozen band, stated here once, and relied on nowhere else in the paper.}

\section{Discussion}
\label{sec:discussion}

\subsection{What is now measured}

On the confirmation side, four coupled-regime regularities are now measured properties of a real frontier-LLM economy, at millinat-scale precision, under frozen bands: relative growth equals relative claimed information (46 millinats worst-case); coalition value is submodular where channels are conditionally independent and supermodular by 90\% of the ideal bit where they are synergistic; joint growth respects the entropy ceiling exactly; and market selection concentrates wealth on the best-informed agent. None of these follows from single-agent betting theory \citep{kelly1956,barron1988} alone --- each is a statement about the \emph{coupling}. For the ML audience the point is falsifiability: these are laws in nats connecting a measurable epistemic quantity to a measurable economic one, they could have failed at any of five pre-registered bands, and they did not.

Two measured deviations deserve equal prominence. Elicited joint posteriors are \emph{not} the product of elicited singletons even where conditional independence holds (max deviation $0.466$ nats) --- consistency across elicitation contexts is not free, and aggregation schemes that assume it inherit the error. And clone equality fails under absorbing all-in dynamics at default sampling (Section~\ref{sec:p4b}) --- idealized symmetric-agent results can be broken by sampling noise composed with absorbing wealth dynamics, a mechanism likely generic across LLM-agent markets.

\subsection{What the negative result means for models of LLM populations}

The natural modeling assumption for a population of near-rational agents under incentives --- a smooth response surface, locally linear in the levers, supporting marginal analysis --- has no support in either small-model campaign or in any of our 104 frontier runs. Frontier populations in this environment are \emph{discrete attractor systems}: within a basin, behavior is reproducible to three decimal places across seeds; at basin boundaries, the seed decides. The governance-relevant corollary, stated within the scope of this data: \textbf{an intervention on such a population either crosses an attractor boundary or does nothing measurable.} Raising the control incentive by 8\% at B1 achieved nothing; lowering the competing incentive by the same absolute amount achieved everything (\expl\ reading, Section~\ref{sec:p6}; the pre-registered grid supports the step-function structure independently of the arm data). Mean-field linear-response models --- and any regulatory intuition of the form ``a bit more oversight buys a bit more alignment'' --- assume a dispersion that these populations do not hold. Where such assumptions are load-bearing in multi-agent-safety arguments \citep{critch2020,hammond2025}, they should be checked against attractor structure, and interventions evaluated by which basin they land the population in, not by their marginal strength.

We emphasize the epistemic status split: the capacity-region result is a pre-registered \emph{confirmation}; the attractor characterization is a pre-registered \emph{CAP} whose structural content (dispersion collapse, step-function response, bistability) is measured, while its lever-switching interpretation at the P6 points is exploratory and labeled so.

\subsection{Limitations}
\label{sec:limitations}

\textbf{Single vendor, single model family, one frontier tier.} Every verdict-run token came from \texttt{claude-opus-4-8}. Both results are statements about this instrument class until replicated elsewhere; \emph{cross-vendor replication is the open arm of this program}, and the full protocol and analyzer run unchanged against any OpenAI-compatible or similar endpoint. \textbf{Toy scale.} $N = 12$--$20$ agents, $T = 8$--$18$ rounds, 8-state worlds, uniform resource weights; the market has no take, no strategic bet-shading incentive, and no entry/exit. These are the smallest systems in which the predictions are well-posed --- not markets, not deployments. \textbf{Unincentivized elicitation.} Part A's primary protocol asks for stated posteriors; agents are not paid for calibration \citep{gneiting2007}. The live market partially covers the acting-vs-stating gap at small scale --- and the gap law held with $\Ihat$ measured from statements and $\Ghat$ from the same statements' market consequences --- but a fully incentive-compatible elicitation is future work. \textbf{Population-dynamics caveats.} The niche economy's prompt-$\gamma$ calibration is absorbed into the fit constant; the $J$-granularity systematic was pre-stated but is a real $\pm1$--$3$ niche coarseness; the P6 floor definition has the drafting gap named in Section~\ref{sec:p6}; and $\sim$3 runs were double-sampled across a machine migration (cache-replayed; remote data of record; disclosed). \textbf{Amendment history.} Three amendments were committed after the freeze --- each blind, each before the runs it governed, each traceable in the public chain --- and the first Experiment-A pass had real instrument faults that the frozen gates caught (Section~\ref{sec:amendments}). We consider gates-that-fire a feature of the method, and disclose every firing.

\section{Reproducibility statement}
\label{sec:repro}

\ifanon
\else
The repository \url{https://github.com/macrokit/value} contains the frozen pre-registrations and amendments (\texttt{sim/stage2/PREREGISTRATION*.md}), the runners (\texttt{stage2a.py}, \texttt{stage2b.py}, \texttt{gate.py}), the mechanical analyzer (\texttt{analyze\_stage2.py}), the complete raw outputs and frozen analysis (\texttt{results/*.json}), and the commit chain of Section~\ref{sec:chain} establishing that every prediction precedes its data. Running \texttt{python3 sim/stage2/analyze\_stage2.py} reproduces every number in this paper offline, deterministically, at zero cost, from the committed raw outputs; the underlying per-call SQLite cache (19.9\,MB, keyed by full request, with per-call token metering) is committed with the repository (\texttt{sim/stage2/results/gate\_cache.sqlite}), so every individual model response in the experiment is itself auditable. Re-running the experiment from scratch against the live API costs $\approx$ \$139 at current list prices (audit in Appendix~\ref{app:cost}). A mock backend (\texttt{--tier mock}) allows full-protocol dry runs without any API key.
\fi

\section{Conclusion}

In a pre-registered test on frontier LLM agents costing \$138.76, an information-theoretic capacity region for coupled agent economies was confirmed within all five of its frozen primary bands --- relative growth equals relative claimed information to 46 millinats, coalition value bends exactly where and how conditional independence says it must, joint growth saturates the entropy ceiling, and markets select the best-informed --- while the \ifanon\else companion\fi\ mean-field residual-scaling law found no domain to be tested in: every tested LLM population, across three instruments and two capability tiers, either saturates, ignores, or totally collapses the goal dispersion the law assumes, and the frontier population in particular behaves as a discrete attractor system with step-function responses and seed-selected bistability. What an LLM agent knows prices what it earns, to the millinat, the moment agents are coupled through a market. How an LLM \emph{population} responds to incentives is not smooth at all: interventions switch attractors, or they do nothing.

\ifanon\else
\subsubsection*{Acknowledgments}
The experiment was self-funded (\$138.76 in metered API spend). No external compute, grants, or institutional resources were used.
\fi

\bibliography{refs}
\bibliographystyle{tmlr}

\appendix

\section{Frozen bands and decision rule}
\label{app:bands}

Verbatim summary of \texttt{sim/stage2/PREREGISTRATION.md} \S4 (commit \texttt{4c0a7a1}, which precedes all Experiment A/B runs).

\paragraph{Prediction (i).}
\begin{center}
\small
\begin{tabular}{@{}lll@{}}
\toprule
ID & Prediction & Band \\
\midrule
P1 & parimutuel $\Ghat_a - \Ghat_b = \Ihat_a - \Ihat_b$, all pairs, (a)--(d) & worst deviation $\leq 0.05$ nats \\
P2 & $\Ghat_S$ diminishing returns for (a)--(d); supermodular in (e) & margins $\geq -0.03$; (e) gap $\geq \ln2/2$ \\
P3 & $\Ghat_S \leq H(X)$, all coalitions, all structures & slack $0.02$ \\
P4 & wealth ordered by $\Ihat_a$; clones equal & order in $\geq 4/5$ seeds; clone gap $\leq 10\%$ \\
\bottomrule
\end{tabular}
\end{center}

\paragraph{Prediction (ii).} P5: free log--log fit on in-regime condition means, $p \in [0.6, 1.4]$, $q \in [-1.4, -0.6]$, $R^2 \geq 0.70$, seed-bootstrap 95\% CIs (2000 resamples), signal gate $R_{\mathrm{across}} > 3\sigma_{\mathrm{in}}$ (raw seed std), mean ratio to $Vg/\gamma$ in $[0.33, 3.0]$; regime rule per condition: $V_{\mathrm{ss}} \geq 5.31$, $\bar k_{\mathrm{ss}} \leq 10$, above-peak $\leq 0.20$, parse $\geq 0.90$, residual $> 0$; the uniformity clause ($V_{\mathrm{ss}}$ within $\pm 25\%$ of $21.25$ and $\bar k$ within $\pm 1$ of $7.5$) names a condition non-responsive; $\geq 6$ in-regime conditions spanning $\geq 4\times$ required, else CAP. P6: at B1 $(6, 0.7)$ and B2 $(18, 0.7)$ with $\delta = 0.5$: $\mathrm{eff}_g/\mathrm{eff}_\gamma > 1$ at both points, all four arm signals above the floor $\sigma_{\mathrm{in}}\sqrt{2/8}$.

\paragraph{Decision rule (applied mechanically, in order).} (1) PASS if P1--P3 and P5 pass; (2) CLEAN NEGATIVE if the instrument engages but P1/P2/P5 fail with the deviation systematic above the noise bar (alternative-law fit $R^2 \geq 0.70$, $R_{\mathrm{across}} > 3\sigma_{\mathrm{in}}$, leave-one-out stable); (3) CAP otherwise, with the binding gate named. \textbf{Outcome of record: (3) CAP}, named gate ``P5 insufficient in-regime span (0/9; dispersion collapse)'' --- within which P1, P2, P2e, P3, P4a pass their frozen bands, P4b fails (supporting, disclosed), and P6 is formally inconclusive. Under the rule's letter the whole-experiment outcome is CAP because rule (1) requires P5; prediction (i) is confirmed within its own scope, and we report the split exactly as it fell.

\section{Amendment history (full disclosure)}
\label{app:amendments}

\paragraph{Gate Amendment 1 (\texttt{4a8c376}, before any gate run).} The frozen models reject the \texttt{temperature} API parameter; it is omitted everywhere. The provider-default sampling noise (nominally higher than the small-model arms' $0.6$) is the harder-to-pass direction for every concentration-dependent band.

\paragraph{Amendment 2 (\texttt{c79f8fe}, before any Experiment A/B run).} (A2.1) The original \S2 froze naive-Bayes product fusion as \emph{the} coalition rule while \S4's P2e band requires XOR supermodularity $\geq \ln2/2$ --- mathematically unsatisfiable, since a product of uniform singleton posteriors is uniform, giving $\Ghat_{\{1,2\}} = 0$ for \emph{perfect} agents. Corrected: coalition-level elicitation defines $\Ghat_S$; product fusion becomes a pre-registered secondary diagnostic (expected to track (a)--(d) and fail (e) --- both borne out, Section~\ref{sec:xor}). (A2.2) The original live-market structures (a) and (c) give all agents equal channel information, making P4a's ordering degenerate; corrected to the ladder (L) + clones (C) pair with bands unchanged.

\paragraph{Amendment 3 (\texttt{9d685f8}, after the first Experiment-A pass, before the re-run; no Experiment-B call yet).} The frozen \S5 gates fired on parts of the first pass; a cache audit traced each firing to an instrument fault. (A3.1) The analyzer recorded elicitation-gate failures but did not exclude affected structures from the verdicts, contrary to the frozen text; fixed. (A3.2) Market output protocol: 36\% of first-pass bet outputs were parseable (agents reasoned past the 600-token ceiling before emitting JSON; the 64\% uniform fallbacks mechanically produced the observed wealth dynamics). Corrected to a JSON-first instruction, a 900-token ceiling, one pre-stated format-only retry per call, and an explicit market parse gate at $0.90$. The structure-(e) coalition prompt contained a self-contradiction (an instance told it ``does NOT see'' a signal the coalition prompt then showed); coalition prompts now compose observations neutrally. First-pass (e) posteriors are disclosed as contaminated, evidence of nothing. (A3.3) Scope: structures (b), (d), (e) and the full market re-collected under the corrected protocol; structures (a), (c) passed cleanly under the original protocol and were kept --- passed data is never re-rolled, in either direction (anti-fishing symmetry).

\section{Full per-seed grid data}
\label{app:grid}

Residual $\bar k_{\mathrm{ss}} - k^*$ and dispersion $V_{\mathrm{ss}}$ per seed (seeds 0--7), all 72 grid runs, parse rate $1.000$ in every run. Values re-derive from \texttt{results/stage2b\_raw.json}.

\begin{center}
\small
\begin{tabular}{@{}rrl@{}}
\toprule
$\gamma$ & $g$ & per-seed residuals (top) / $V_{\mathrm{ss}}$ (bottom) \\
\midrule
2 & 0.25 & $[9.94,\ 12.93,\ 13.00,\ 12.00,\ 14.86,\ 0.02,\ 6.97,\ 12.00]$ \\
  &      & $[0.39,\ 0.58,\ 0.00,\ 0.00,\ 1.43,\ 0.10,\ 1.17,\ 0.00]$ \\
\addlinespace
2 & 0.70 & $[9.99,\ 12.00,\ 13.00,\ 12.00,\ 14.60,\ 12.00,\ 9.97,\ 12.00]$ \\
  &      & $[0.04,\ 0.00,\ 0.00,\ 0.00,\ 4.15,\ 0.00,\ 0.25,\ 0.00]$ \\
\addlinespace
2 & 2.00 & $[9.87,\ 12.93,\ 13.00,\ 12.00,\ 14.64,\ 12.00,\ 9.92,\ 12.00]$ \\
  &      & $[0.77,\ 0.67,\ 0.00,\ 0.00,\ 3.11,\ 0.00,\ 1.06,\ 0.00]$ \\
\addlinespace
6 & 0.25 & $[0,\ 0,\ 0,\ 0,\ 0,\ 0,\ 0,\ 0]$ \quad / \quad $[0,\ 0,\ 0,\ 0,\ 0,\ 0,\ 0,\ 0]$ \\
\addlinespace
6 & 0.70 & $[9.96,\ 13.00,\ 13.00,\ 12.00,\ 14.55,\ 0.25,\ 9.95,\ 12.00]$ \\
  &      & $[0.21,\ 0.00,\ 0.00,\ 0.00,\ 4.85,\ 2.36,\ 0.29,\ 0.00]$ \\
\addlinespace
6 & 2.00 & $[9.90,\ 12.00,\ 12.95,\ 12.00,\ 14.57,\ 12.00,\ 9.90,\ 12.00]$ \\
  &      & $[0.89,\ 0.00,\ 0.32,\ 0.00,\ 4.08,\ 0.00,\ 0.57,\ 0.00]$ \\
\addlinespace
18 & 0.25 & $[0,\ 0,\ 0,\ 0,\ 0,\ 0,\ 0,\ 0]$ \quad / \quad $[0,\ 0,\ 0,\ 0,\ 0,\ 0,\ 0,\ 0]$ \\
\addlinespace
18 & 0.70 & $[0,\ 0,\ 0,\ 0,\ 0,\ 0,\ 0,\ 0]$ \quad / \quad $[0,\ 0,\ 0,\ 0,\ 0,\ 0,\ 0,\ 0]$ \\
\addlinespace
18 & 2.00 & $[9.90,\ 12.83,\ 12.93,\ 12.00,\ 14.61,\ 0.45,\ 9.78,\ 12.00]$ \\
  &      & $[0.43,\ 1.60,\ 0.35,\ 0.00,\ 3.57,\ 4.00,\ 1.37,\ 0.00]$ \\
\bottomrule
\end{tabular}
\end{center}

P6 direction arms (8 seeds each): B1 baseline $(\gamma, g) = (6, 0.7)$: mean residual $10.589$; arm $\gamma \to 6.5$: $12.078$ (per-seed $[9.91, 12.86, 13.00, 12.00, 14.89, 12.00, 9.97, 12.00]$); arm $g \to 0.2$: $0.000$ (all seeds). B2 baseline $(18, 0.7)$: $0.000$; both arms $0.000$ (all seeds). Analyzer efficacies: B1 $\mathrm{eff}_\gamma = -2.978$, $\mathrm{eff}_g = +21.178$, ratio $-7.11$; B2 $0/0$ (ratio undefined). With $\sigma_{\mathrm{in}}$ undefined (no in-regime condition), no arm is formally resolvable: P6 = INCONCLUSIVE (Section~\ref{sec:p6}).

\section{Elicited information and coalition throughputs}
\label{app:elicitation}

Claimed information $\Ihat_a$ per agent (nats), from 12-repeat averaged elicited posteriors, exact enumeration ($124$ cells total; every cell $\geq 6/12$ surviving repeats in the verdict pass):

\begin{center}
\small
\begin{tabular}{@{}lllll@{}}
\toprule
Structure & Agent 0 & Agent 1 & Agent 2 & exact $I$ \\
\midrule
(a) disjoint & 0.6931 & 0.6931 & 0.6931 & 0.6931 \\
(b) overlap & 1.3009 & 1.3863 & 1.3863 & 1.3863 \\
(c) clones & 0.6931 & 0.6931 & 0.6931 & 0.6931 \\
(d) noisy & 0.3681 & 0.3681 & 0.3681 & 0.3681 \\
(e) XOR & 0.0000 & 0.0000 & --- & 0.0000 \\
\bottomrule
\end{tabular}
\end{center}

Selected coalition throughputs $\Ghat_S$ (coalition-level elicitation) vs exact $I(X; Y_S)$, in nats:

\begin{center}
\small
\begin{tabular}{@{}llll@{}}
\toprule
Structure & Coalition & $\Ghat_S$ & $I(X;Y_S)$ \\
\midrule
(a) disjoint & $\{0,1\}$ & 1.3863 & 1.3863 \\
(a) disjoint & $\{0,1,2\}$ & 2.0794 & 2.0794 \\
(b) overlap & $\{0,1\}$ & 2.0794 & 2.0794 \\
(b) overlap & $\{0,1,2\}$ & 1.7688 & 2.0794 \\
(c) clones & $\{0,1\}$ (clone pair) & 0.6931 & 0.6931 \\
(c) clones & $\{0,1,2\}$ & 1.3863 & 1.3863 \\
(d) noisy & $\{0,1,2\}$ & 0.6383 & 1.1042 \\
(e) XOR & $\{0\}$, $\{1\}$ & 0.0000 & 0.0000 \\
(e) XOR & $\{0,1\}$ & 0.6212 & 0.6931 \\
\bottomrule
\end{tabular}
\end{center}

Note $\Ghat_S \leq I(X;Y_S)$ throughout (elicitation can only lose information; P3's ceiling then follows a fortiori), with the clone pair adding exactly zero and the XOR pair recovering 90\% of the joint bit. The (b) triple and (d) coalitions under-recover their exact $I$ --- elicited joint posteriors lose information on redundant/noisy composites --- while remaining fully consistent with P2's submodularity bands and P3's ceiling. Live-market terminal wealth shares (ladder, 5 seeds): clean agent $[0.999, 1.000, 1.000, 1.000, 1.000]$; clones: $[(0.75, 0.25), (0.375, 0.625), (0.5, 0.5), (0.5, 0.5), (\mathrm{nan})]$ (Section~\ref{sec:p4b}).

\section{Cost audit}
\label{app:cost}

Audited from the metered call cache (\texttt{results/gate\_cache.sqlite}, \texttt{usage} table), at the calibrated list prices frozen in the pre-registration (\$5/M input, \$25/M output for the frontier tier):

\begin{center}
\small
\begin{tabular}{@{}lrrrr@{}}
\toprule
Tier & Calls & Tokens in & Tokens out & Cost \\
\midrule
\texttt{claude-opus-4-8} (gate + Experiments A, B) & 43{,}074 & 16{,}674{,}507 & 2{,}126{,}135 & \$136.53 \\
\texttt{claude-sonnet-5} (protocol debug only) & 1{,}634 & 632{,}337 & 22{,}291 & \$2.23 \\
\midrule
\textbf{Total (hard cap \$160, pre-call abort)} & 44{,}708 & 17.3M & 2.15M & \textbf{\$138.76} \\
\bottomrule
\end{tabular}
\end{center}

The capability gate alone cost \$8.76 (of its own \$30 cap). The frozen spend order (Experiment A first, then the B grid, then the direction arms) was designed so that a budget stop would preserve the most complete publishable unit; the cap was never hit.

\end{document}